\documentclass[10pt,twocolumn,letterpaper]{article}

\usepackage{cvpr}              %
\usepackage{amsmath}
\usepackage{amsfonts}
\usepackage{bm}  
\usepackage{graphicx}
\usepackage{tabularx}
\usepackage{multirow} 

\usepackage[dvipsnames]{xcolor}

\definecolor{cvprblue}{rgb}{0.21,0.49,0.74}
\usepackage[pagebackref,breaklinks,colorlinks,citecolor=cvprblue]{hyperref}

\def\paperID{8545} %
\def\confName{CVPR}
\def\confYear{2024}

\begin{document}

\title{UDiFF: Generating Conditional Unsigned Distance Fields with Optimal Wavelet Diffusion}  

\author{Junsheng Zhou$^{1*}$, Weiqi Zhang$^{1*}$, Baorui Ma$^{1,2\dag}$, Kanle Shi$^3$, Yu-Shen Liu$^{1\dag}$, Zhizhong Han$^4$\\
School of Software, Tsinghua University, Beijing, China$^1$\\
BAAI, Beijing, China$^2$, Kuaishou Technology, Beijing, China$^3$\\
Department of Computer Science, Wayne State University, Detroit, USA$^4$\\
{\tt\small zhoujs21@mails.tsinghua.edu.cn, zwq23@mails.tsinghua.edu.cn, brma@baai.ac.cn}\\
{\tt\small shikanle@kuaishou.com, liuyushen@tsinghua.edu.cn,\hspace {3mm}h312h@wayne.edu}
}

\twocolumn[{%
\renewcommand\twocolumn[1][]{#1}%
\maketitle
\begin{center}
\centering
\captionsetup{type=figure}
\vspace{-7mm}

\includegraphics[width=\linewidth]{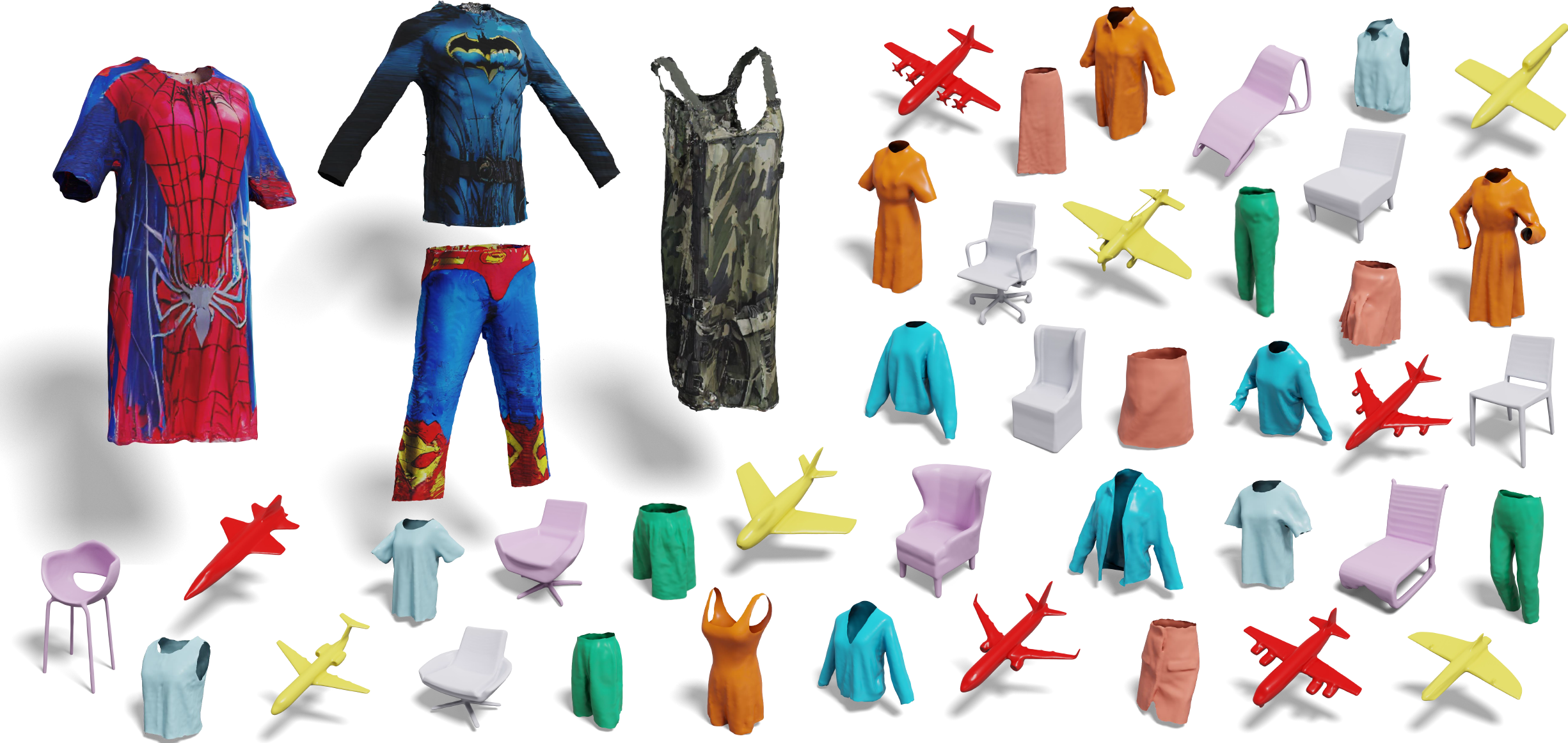}
\captionof{figure}{
Diverse shapes with and without open surfaces generated by our UDiFF model. \textbf{Top-Left:} Conditional generation of clothes with prompts `A short-sleeved dress in spiderman style', `A Batman upper with long sleeves', `A superman pant', `A camouflage slip dress'. \textbf{Around:} A shape gallery generated by UDiFF conditionally and unconditionally.
}
\label{fig:teaser}
\end{center}%
}]

\if TT\insert\footins{\footnotesize{
*Equal contribution. $\dag$ Corresponding authors. This work was supported by National Key R\&D Program of China (2022YFC3800600), the National Natural Science Foundation of China (62272263, 62072268), and in part by Tsinghua-Kuaishou Institute of Future Media Data.}}\fi

\begin{abstract}
Diffusion models have shown remarkable results for image generation, editing and inpainting. Recent works explore diffusion models for 3D shape generation with neural implicit functions, i.e., signed distance function and occupancy function. However, they are limited to shapes with closed surfaces, which prevents them from generating diverse 3D real-world contents containing open surfaces. 
In this work, we present UDiFF, a 3D diffusion model for unsigned distance fields (UDFs) which is capable to generate textured 3D shapes with open surfaces from text conditions or unconditionally. Our key idea is to generate UDFs in spatial-frequency domain with an optimal wavelet transformation, which produces a compact representation space for UDF generation. Specifically, instead of selecting an appropriate wavelet transformation which requires expensive manual efforts and still leads to large information loss, we propose a data-driven approach to learn the optimal wavelet transformation for UDFs. We evaluate UDiFF to show our advantages by numerical and visual comparisons with the latest methods on widely used benchmarks. Page: \url{https://weiqi-zhang.github.io/UDiFF}.

\end{abstract}

\section{Introduction}

Probabilistic diffusion models \cite{sohl2015deep, ho2020denoising} have largely revolutionized 2D content generation. Recent advancements, such as DALL-E 2 \cite{ramesh2022hierarchical} and Stable Diffusion \cite{rombach2022high},  have been widely used in text-to-image generation, image inpainting, etc. A series of works \cite{luo2021diffusion, smith2017improved} try to replicate these success in 3D content generation by developing diffusion models for point clouds or voxels, but fails to produce high fidelity results due to the limited resolution in voxels and the discreteness of points. Recent approaches \cite{hui2022neural, chou2023diffusion, zhang20233dshape2vecset} explore diffusion models to generate 3D shapes as neural implicit functions, i.e., signed distance function (SDF) \cite{park2019deepsdf,ma2021neural} and occupancy function (Occ) \cite{mescheder2019occupancy}. However, they are limited to generate closed shapes since both SDF and Occ model the internal and external relations of 3D locations for representing 3D shapes. This prevents previous 3D implicit diffusion models from generating diverse 3D real-world contents containing open surfaces. 

Another challenge in diffusion-based 3D generative models is how to define a compressing transform schema for achieving compact implicit representations which can be learned by diffusion models efficiently. Some works train a variational auto-encoder (VAE) \cite{kingma2013auto} for converting shapes into triplane \cite{shue20233d, gupta20233dgen} or single latents \cite{nam20223d} for latent diffusion. However, the relative limited 3D data makes it difficult to train a stable VAE. Instead, another series of works (e.g. WaveGen \cite{hui2022neural}) seek to leverage explicit transform in another domain (e.g. wavelet transform \cite{daubechies1990wavelet}) for direct compression. Nevertheless, they need to select an appropriate wavelet type, which demands extensive manual efforts and can still result in significant information loss during wavelet recovery.

To address these issues, we propose UDiFF, a 3D diffusion model for unsigned distance fields 
\cite{chibane2020neural, Zhou2022CAP-UDF} which is capable of generating textured 3D shapes without geometric limits on the surface watertightness (e.g. contain open surfaces). Compared to commonly-used SDF or Occ, UDF has proven to be an advanced representation that supports arbitrary typologies and remain strong generalization. Going beyond pure unconditioned models, we incorporate conditions achieved from CLIP \cite{radford2021learning} models to UDiFF by introducing conditional cross-attentions. This enables to control 3D generation using the text and image signals.  Previous works merely focus on generating geometries which lead to a lack of appearance and prevent them from creating diverse and visual-appealing 3D models, while we get inspiration from Text2Tex \cite{chen2023text2tex} to simultaneously generate textures for universal 3D content creation.

Adapting existing SDF-based diffusion models directly to UDF does not work well.
The difficulty arises from the significantly greater complexity of UDF compared to SDF, particularly in the context of the non-differential zero-level set.
To solve this issue, we introduce UDiFF as a diffusion model in the spatial-frequency domain based on an optimal wavelet transformation, which produces a compact representation space for UDF generation. Instead of engaging in selecting a suitable wavelet transformation, which is tedious and often results in significant information loss, we employ a data-driven approach to obtain an optimal wavelet filter for representing UDFs. We minimize the \mbox{unsigned} distance errors during a self-reconstruction through  the wavelet transformation, especially near the zero-level set of UDFs. This preserves the geometry details during wavelet transformation, which leads to the high-fidelity generation of 3D geometries. We evaluate UDiFF for generating 3D shapes with open surfaces and closed surfaces using conditions or unconditionally under DeepFashion3D \cite{zhu2020deep}  and ShapeNet \cite{chang2015shapenet} datasets. The experimental results demonstrate that UDiFF achieves promising generation performance compared to the existing state-of-the-art approaches, in both qualitative and quantitative evaluations.
Our main contributions can be summarized as follows.
\begin{itemize}
    \item We propose UDiFF, a 3D diffusion model for unsigned distance fields which is capable of generating real-world textured 3D shapes with open surfaces from text conditions or unconditionally.
    \item We introduce an optimal wavelet transformation for UDF  through data-driven optimization, and justify that the spatial-frequency domain learned through this transformation is a compact domain suitable for UDF generation. 
    \item We evaluate UDiFF for generating 3D shapes with both open and closed surfaces, and show our superiority over the state-of-the-art methods.
\end{itemize}

\begin{figure*}[!t]
  \centering
  \includegraphics[width= \textwidth]{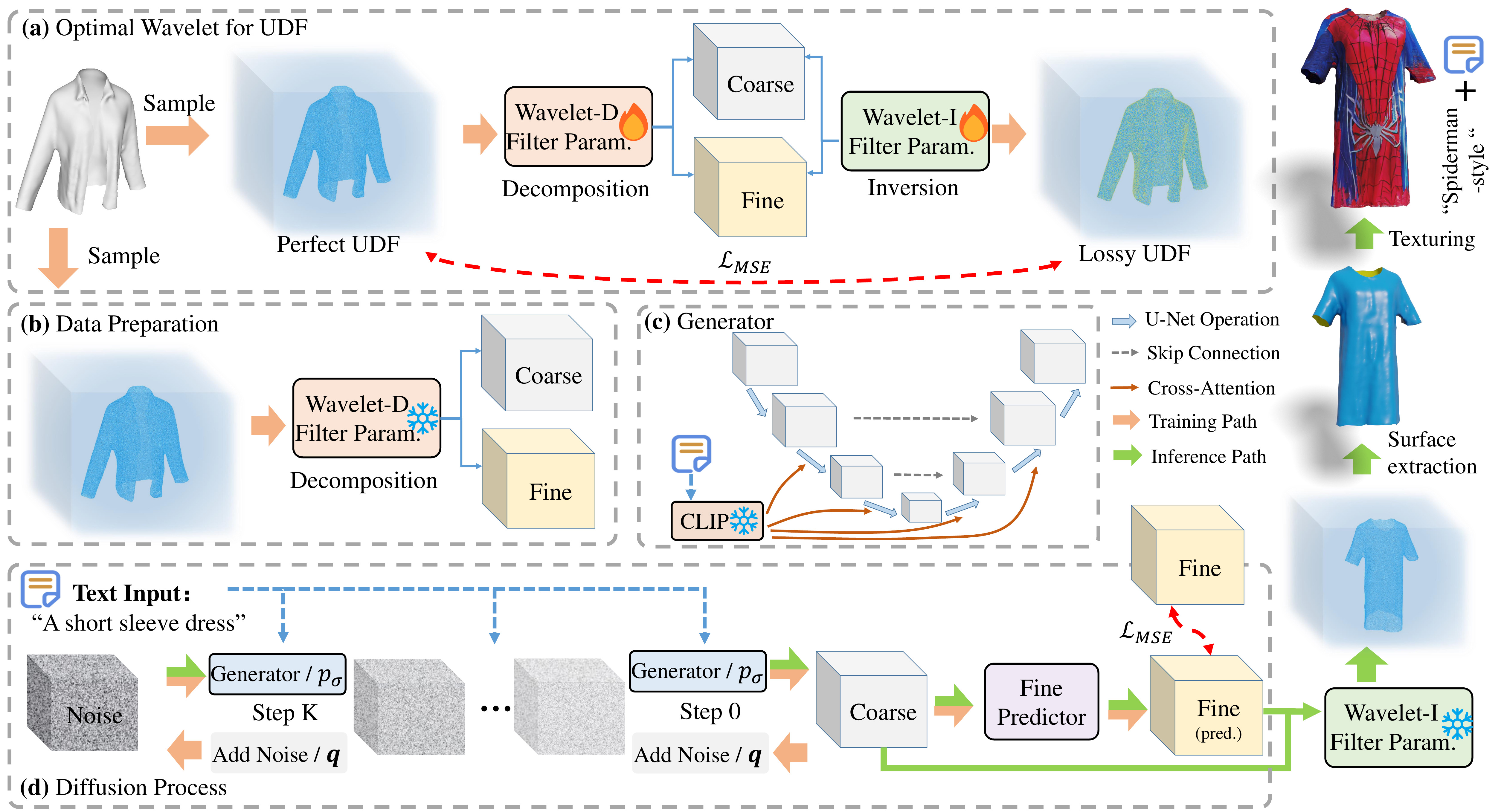} 
  \vspace{-0.6cm}
  
  \caption{\textbf{Overview of UDiFF. (a)} We propose a data-driven approach to attain the optimal wavelet transformation for UDF generation. We optimize wavelet filter parameters through the decomposition and inversion by minimizing errors in UDF self-reconstruction. \textbf{(b)} We fix the learned decomposition wavelet parameters and leverage it to prepare the data as a compact representation of UDFs including pairs of coarse and fine coefficient volumes. \textbf{(c) }is the architecture of the generator in diffusion models, where text conditions are introduced with cross-attentions. \textbf{(d) }The diffusion process of UDiFF. We train the generator to produce coarse coefficient volumes from random noises guided by input texts and train the fine predictor to predict fine coefficient volumes from the coarse ones. Follow the \textcolor[RGB]{0,128,0}{\textbf{green}} arrows for inference, we start from a random noise and an input text to leverage the trained generator to produce a coarse coefficient volume. The trained fine predictor then predicts the fine coefficient volume. Together with the coarse one, we recover the UDFs with the fixed pre-optimized inversion wavelet filter parameters. Finally, we extract surfaces from UDFs and further texture them with the guiding text.}
  \label{fig:overview}
  \vspace{-0.4cm}
\end{figure*}

\section{Related Work}
With the rapid development of deep learning, the neural networks have shown great potential in 3D applications \cite{xiang2022SPD, ma2023towards, wen20223d, zhang2023fast, Zhou2023VP2P, jin2023multi, zhou20223d, huang2023neusurf, li2023learning, ma2023geodream, wen2022pmp, zhou2023uni3d, zhou2022self, li2022neaf}. We mainly focus on learning generative Neural Implicit Functions with networks for generating 3D shapes.

\subsection{Neural Implicit Representations}
Recently, Neural Implicit Functions (NIFs) have shown promising results in surface reconstruction \cite{park2019deepsdf, mescheder2019occupancy, BaoruiNoise2NoiseMapping}, novel view synthesis \cite{mildenhall2020nerf, muller2022instant}, image super-resolution \cite{sitzmann2020implicit, brooks2023instructpix2pix}, etc. The NIFs approaches train a neural network to represent shapes and scenes with signed distance functions (SDFs) \cite{park2019deepsdf, chougensdf} or binary occupancy \cite{mescheder2019occupancy, peng2020convolutional}, where the marching cubes algorithm \cite{lorensen1987marching} is then used to extract surfaces from the learned NIFs. OccNet and DeepSDF \cite{mescheder2019occupancy, park2019deepsdf} are the pioneers of NIFs which learn global latent codes for representing 3D shapes with MLP-based decoder to achieve occupancies or signed distances. The subsequent approaches \cite{peng2020convolutional, jiang2020local} leverage more latent codes to represent detailed local geometries. PCP \cite{PredictiveContextPriors} and OnSurf \cite{On-SurfacePriors} introduce predictive context priors and on-surface prior to enhance the representation ability of NIFs.

Occupancy and SDFs are mainly suitable to represent closed shapes. Recent works explore the neural unsigned distances (UDFs) \cite{chibane2020neural, Zhou2022CAP-UDF, chen20223psdf, wanghsdf, long2022neuraludf, liu2023neudf, zhou2023levelset} to represent shapes and scenes with open surfaces. NDF \cite{chibane2020neural} designs a hierarchical neural network to learn UDFs with ground truth distance supervisions. GIFS \cite{ye2022gifs} learns UDFs and represents shapes with query relationships. CAP-UDF \cite{Zhou2022CAP-UDF} and LevelSetUDF \cite{zhou2023levelset} propose consistency-aware constraints and level set projections to stabilize the optimization of UDFs and produce more accurate geometries.

\subsection{Diffusion-based 3D Generative Models}
Generating 3D contents plays the key role in augmented/virtual reality and has been widely explored in the past few years. Earlier works transfer the success of GAN \cite{goodfellow2020generative}, VAE \cite{kingma2013auto} and the flow-based model \cite{kingma2018glow} in image generation to the 3D domain for generating 3D shapes represented as point clouds \cite{hui2020progressive,li2021sp,cai2020learning,yang2019pointflow,nichol2022point} and voxels \cite{smith2017improved, wu2016learning}. PointDiff \cite{luo2021diffusion} introduces the powerful diffusion models for point cloud generation. Some advanced works \cite{zhou20213d, zeng2022lion} combining the voxel and point representations were proposed for more robust 3D generation with diffusion models.

More recently, some approaches \cite{gao2022get3d,hui2022neural,li2023diffusion,cheng2023sdfusion,shue20233d} try to combine the diffusion models and neural implicit representations for generating high-quality 3D shapes. These methods generate signed distance fields \cite{hui2022neural,li2023diffusion,gupta20233dgen,koo2023salad,chou2023diffusion} or occupancy fields \cite{zhang20233dshape2vecset} with diffusion models and extract the meshes from the fields with the marching cubes \cite{lorensen1987marching}. For the efficient training of diffusion models, methods like Diffusion-SDF \cite{chou2023diffusion} and 3D-LDM \cite{nam20223d} train a VAE for converting shapes into latent codes for latent diffusion. But the relative small number of 3D samples for training makes it difficult to train a stable VAE. WaveGen \cite{hui2022neural} was proposed to explicitly compress SDFs in frequency domain with wavelet transform, but it is limited to the information loss during the wavelet recovery.

The advances in NIFs-based 3D generative models have shown significant improvements in the generation qualities, however, they are limited to generate closed surfaces. %
This prevents them from generating diverse 3D contents in real world. In this work, we focus on generating UDFs for open surfaces with textures using a 3D diffusion model.

\section{Method}

\noindent\textbf{Overview.}
The overview of UDiFF is shown in Fig. \ref{fig:overview}. UDiFF is a 3D generative model which takes texts as conditions and generates general textured 3D shapes with either open or closed surfaces. We will start by introducing the novel approach to obtain an optimal wavelet transform for a compact UDF representation and the data preparation process for training diffusion models in Sec. \ref{sec.3.1}. We then present the designed conditional diffusion framework for UDF generation and the generator network in Sec. \ref{sec.3.2}. Finally, we extract surfaces from the generated UDF and further add textures on the mesh with the guiding text in Sec.~\ref{sec.3.3}.

\subsection{Optimal Wavelet Transformation for UDFs}
\label{sec.3.1}
One main challenge in diffusion-based 3D generative models is to search for a compact representation space for diffusion model to learn efficiently.
WaveGen \cite{hui2022neural} adopts an explicit wavelet transform on the SDF volumes ($256^3$) to decompose them into coarse coefficient volumes and fine coefficient volumes with much lower resolutions. The naive wavelet transform leads to large information loss since the manually selected wavelet is not capable of representing various shapes as accurate distance functions.

To represent UDFs in a compact way, we follow WaveGen to adopt multi-scale wavelet transform \cite{daubechies1990wavelet, mallat1989theory} as the compressing schema, keeping only the coefficients at a relative small scale of $\mathcal{J}=3$ for efficient shape learning. However, the UDF is significantly more complex and unstable than SDF, particularly in the area of non-differential zero-level sets, where the geometry details that the wavelet compressing does not preserve will severely affect the generation of UDFs. Thus, a suitable wavelet filter with much less information loss but remains compact and efficient for UDFs is vital. 

\begin{figure}[!t]
  \centering
  \includegraphics[width= \linewidth]{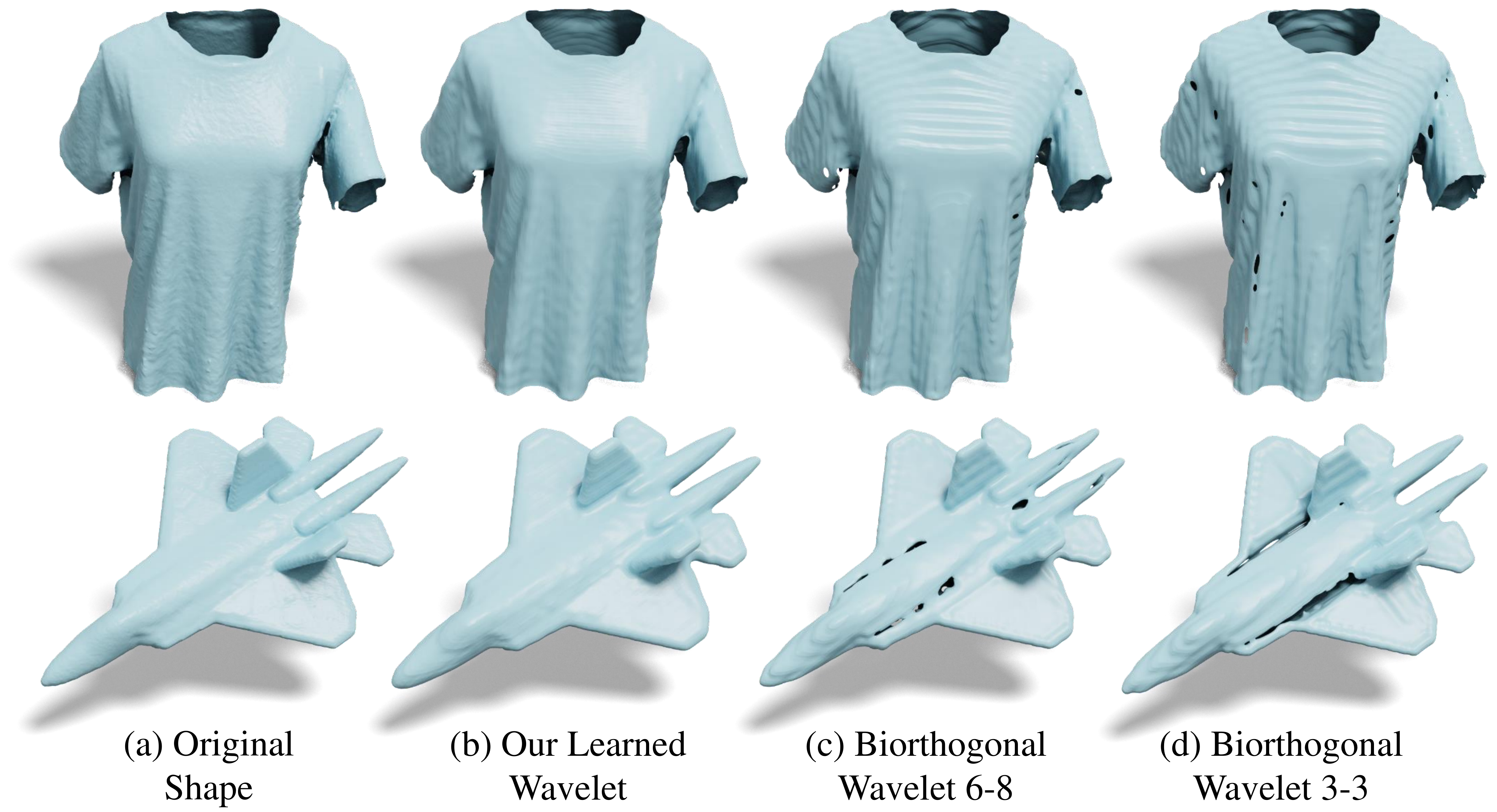} 
  \vspace{-0.3cm}
  \caption{\textbf{Comparisons of reconstructions with different wavelet filters. } (a) The input shapes from DeepFashion3D \cite{zhu2020deep} and ShapeNet \cite{chang2015shapenet}, from where we sample UDFs to prepare compact wavelet representations. (b) The surfaces extracted from the recovered UDF with decomposition and inversion by our learned wavelet filter. (c,d) The surfaces extracted from the recovered UDF with manual chosen wavelet filters.  }
  \label{fig:filter}
  \vspace{-0.3cm}
  
\end{figure}

To this end, instead of manually searching for the appropriate wavelet filter which demands costly efforts and is still hard to reduce the information loss, we propose a data-driven approach to learn the optimal wavelet filter parameters for UDFs through learning-based optimization as shown in Fig.~\ref{fig:overview}(a). Specifically, we define a learnable biorthogonal wavelet filter which consists of a decomposition filter $\phi^D_{\theta}$ and an inversion filter $\phi^I_{\delta}$ with learnable filter parameters $\theta$ and $\delta$. Given a set of shapes $\{S_i\}_{i=1}^N$, we first sample the UDF volume $U_i$ for each shape at a resolution of $256^3$ and truncate the distance values in $U_i$ to [0, 0.1], and then compress it into a coarse coefficient volume and a fine coefficient volume with the learnable decomposition filter $\phi^D_{\theta}$ as:
\begin{equation}
    \{C_i, F_i\} = \phi^D_{\theta}(U_i).
    \label{eq.decom}
\end{equation}
We then predict the lossy UDF $\hat{U}$ from $C_i$ and $F_i$ with the learnable inversion filter $\phi^I_{\delta}$ as:

\begin{equation}
    \{\hat{U_i}\} = \phi^I_{\delta}(C_i, F_i).
    \label{eq.inv}
\end{equation}

The target is to optimize the filter parameters $\theta$ and $\delta$ by minimizing the information loss during wavelet decomposition and inversion, formulated as:
\begin{equation}
    \min_{\theta,\delta} \sum_{i=1}^{N}  \mathcal{L}_{\mathrm{MSE}} (w_i^{\gamma} \hat{U_i}, w_i^{\gamma} U_i).
\end{equation}
where $w_i^{\gamma}$ is the weights for enforcing the optimization to focus on the space near the zero-level set of UDF.  $w_i^{\gamma}$ has the same size as $U_i$ for weighting each grid in the UDF volume, where we define $w_i^{\gamma}$ according to a threshold $\gamma$ to mask the grids with distances larger than $\gamma$.

After data-driven optimization of the wavelet filters $\phi^D_{\theta}$ and $\phi^I_{\delta}$, we learn the optimal wavelet transform with much less information loss and can faithfully reconstruct the original UDF while remains compact. We show the comparison on the wavelet filters in Fig.~\ref{fig:filter}, where the surfaces reconstructed from UDF with our learned wavelet filter in Fig.~\ref{fig:filter}(b) are much smoother and more accurate than the reconstructions with common filters like Biorthogonal wavelet 3-3 in Fig.~\ref{fig:filter} (d). Specifically, Biorthogonal wavelet 6-8 in Fig.~\ref{fig:filter} (c) is the carefully chosen filter by WaveGen from a series of wavelet filters, where our learned filter significantly outperforms the manually selected filters in compressing and recovering UDF. The reason is that the filters learned by data-driven optimizing from UDF datasets are much more suitable to specific characters of UDFs, which preserves more geometry details. 

With the learned optimal wavelet filter, we then leverage it to represent UDFs as a compact representation for training diffusion models. As shown in Fig.~\ref{fig:overview} (b), we fix the parameters for $\phi^D_{\theta}$ and produce the paired coarse efficient $\{C_i\}_{i=1}^M$ volumes and fine efficient volumes $\{F_i\}_{i=1}^M$ by decomposing $U_i$ with Eq.~(\ref{eq.decom}).

\subsection{Conditional UDF Diffusion}
\label{sec.3.2}
\noindent\textbf{Generator Architecture.} We first introduce the network details of diffusion generators for 3D volumes, as shown in Fig.~\ref{fig:overview}(c). The generator shares a similar U-Net architecture as Stable-Diffusion \cite{rombach2022high, nichol2021improved}, where the 2D convolutions are replaced with 3D ones for handling 3D volumes. Each U-Net operation in Fig. \ref{fig:overview}(c) contains $3\times3\times3$ residual blocks, pooling layers and down/up-sampling layers. For introducing text conditions to diffusion models, we first encode the input texts with frozen CLIP \cite{radford2021learning} models to produce text embeddings and then fuse them into the volume features with cross-attention layers. 
 
\noindent\textbf{Learning Diffusion Models.} We develop our 3D generative model UDiFF based on diffusion probabilistic models \cite{sohl2015deep, ho2020denoising}. The diffusion process is to generate coarse coefficient volumes which represents the general geometry of 3D shapes from random volume noises, as shown in Fig.~\ref{fig:overview} (d). We define $\{\mathcal{C}_0, \mathcal{C}_1, ...,\mathcal{C}_T\}$ as the forward process $q(\mathcal{C}_{0:T})$ which gradually transforms a real data $\mathcal{C}_0$ into Gaussian noise ($\mathcal{C}_T$) by adding noises, where $\mathcal{C}_0$ is a sample from the coarse coefficient data $\{C_i\}_{i=1}^M$. The diffusion backward process $p_\sigma(\mathcal{C}_{0:T})$ leverages the generator with parameter $\sigma$ to denoise $\mathcal{C}_T$ into a real data sample. The learning schema is to train the generator to maximize the generation probability of the target, i.e. $p_\sigma(\mathcal{C}_{0})$. We follow DDPM \cite{ho2020denoising} to simplify the optimization target to predict noises $\epsilon_{\sigma}$ with the generator, formulated as:
\begin{equation}
    \label{eq.dmloss}
    \min_{\sigma} \mathbb{E}_{\mathcal{C}_0, t, {\epsilon}\sim\mathcal{N}(0,1)}\left[\left\|{\epsilon}-{\epsilon}_{\sigma}\left({\mathcal{C}}_t, t\right)\right\|^{2}\right],
\end{equation}
where $t$ is a time step and $\epsilon$ is a noise volume sampled from the unit Gaussian distribution $\mathcal{N}$. 

\noindent\textbf{Condition-Guided 3D Diffusion.} Up to this point, we have covered the generative diffusion process without conditions. For a controllable generation of unsigned distance fields, we further introduce a conditioning mechanism \cite{rombach2022high} into the diffusion process by cross-attention. Specifically, given an input text $y$, we first leverage a frozen CLIP text encoder $\tau$ to project $y$ into the condition embedding $\tau(y)$. The embedding is then fused into the U-Net layers of generator with cross attention modules implemented as $\text{Attention}(Q, K, V) = \text{softmax}\left(\frac{QK^T}{\sqrt{d}}\right) \cdot V$, where $Q = W^{(i)}_Q \cdot  \varphi_i(\mathcal{C}_t), \; K = W^{(i)}_K \cdot \tau(y)$ and $V = W^{(i)}_V \cdot \tau(y) .$ Here, $\varphi_i(\mathcal{C}_t)$ is the output of an intermediate layer of the U-Net and $W^{(i)}_V$, $W^{(i)}_Q$ \& $W^{(i)}_K$ are learnable matrices. 

The cross-attention mechanism learns a mapping from the input text condition to the coefficient volumes which represent the geometric generations. The optimizing target in Eq. (\ref{eq.dmloss}) is then modified as:
\begin{equation}
    \min_{\sigma} \mathbb{E}_{{\mathcal{C}_0}, y, t, {\epsilon}\sim\mathcal{N}(0,1)}\left[\left\|{\epsilon}-{\epsilon}_{\sigma}\left({\mathcal{C}}_t, \tau(y), t\right)\right\|^{2}\right],
\end{equation}

\noindent\textbf{Fine Predictor.} The last module for learning to generate UDFs is the fine predictor $f$ which predicts fine coefficient volumes from the generated coarse ones. We follow WaveGen \cite{hui2022neural} to implement $f$ with the similar U-Net architecture as the generator. We train $f$ with pairs of coarse and fine coefficient volumes $\{C_i, F_i\}$ with MSE loss to minimize the differences between $F_i$ and the prediction $f(C_i)$.

\subsection{Generating Novel 3D Shapes}
\noindent\textbf{Generating UDFs at Inference.} With the learned optimal wavelet filters and the trained conditional diffusion models, we can now generate novel 3D shapes as shown in the green arrows in Fig.~\ref{fig:overview}. Starting from a random volume noise and an input text $y$, we leverage the generator to produce a coarse coefficient $C'$ volume by removing noises iteratively with the guidance of $y$. The fine predictor then predicts the fine coefficient volume $F'$, together with $C'$ to generate the UDF $U'$ by wavelet inversion with the learned filter $\phi^I_{\delta}$ as Eq.~(\ref{eq.inv}).

\noindent\textbf{Surface Extraction and Texturing}
After generating a novel UDF $U'$, we extract the zero-level set of $U'$ as a surface. The recent works \cite{guillard2021meshudf, Zhou2022CAP-UDF} leverage the gradients at UDF as the signals to mesh UDFs, however, the approximated gradients of generated UDF may not be stable enough at the zero-level set, which leads to errors and holes. We therefore adopt DCUDF \cite{hou2023robust} with double covering to mesh the generated UDF of UDiFF. Please refer to the supplementary for the adaptions to DCUDF.
Finally, to create visual-appealing 3D models, we drew inspiration from Text2Tex \cite{chen2023text2tex}. This helps to generates textures for the extracted mesh while leveraging the text guidance within a progressive rendering-based texturing framework.

\label{sec.3.3}

\section{Experiment}

\begin{table}[t!]
\centering
\newcolumntype{Y}{>{\centering\arraybackslash}X}
\caption{\textbf{Quantitative comparison of shape generation under DeepFashion3D dataset.} 
MMD-CD scores and MMD-EMD scores are scaled by $10^3$ and $10^2$, respectively. }
\vspace{-0.25cm}
\footnotesize
{
\setlength{\tabcolsep}{0.2em}
\renewcommand{\arraystretch}{1.0}
\definecolor{LightCyan}{rgb}{0.88,1,1}
\definecolor{Gray}{gray}{0.85}
\begin{tabularx}{\linewidth}{>{\centering}m{2.5cm}| Y Y Y Y Y Y }
  \toprule
  \multirow{2}{*}{Method}
   & \multicolumn{2}{c}{COV $\uparrow$} & \multicolumn{2}{c}{MMD $\downarrow$} & \multicolumn{2}{c}{1-NNA $\downarrow$}  \\
    &   CD   &   EMD   &   CD   &   EMD   &   CD   &   EMD   \\
  \midrule
  PointDiff \cite{luo2021diffusion}   & 68.67  & 64.56 & \textbf{11.01} & 15.53 & 83.21  & 87.69  \\
  WaveGen \cite{hui2022neural}  & 62.34 & 51.89 & 15.56  & 17.03 & 92.93 & 94.83  \\
  Diffusion-SDF  \cite{chou2023diffusion}  & 67.09 & 62.03 & 14.79 & 16.63 & 88.98 & 92.63  \\
  LAS-Diffusion \cite{zheng2023locally}  & 67.40 & 56.01 & 14.59 & 16.53 & 88.61 & 91.41  \\
  \midrule
  Ours   & \textbf{69.62} & \textbf{67.72} & 11.60 & \textbf{14.01} & \textbf{81.83} & \textbf{82.14}  \\

  \bottomrule
\end{tabularx}
}
\vspace{-0.5cm}
\label{tab.deepfashion}
\end{table}

\begin{figure}[!t]
  \centering
  \includegraphics[width= \linewidth]{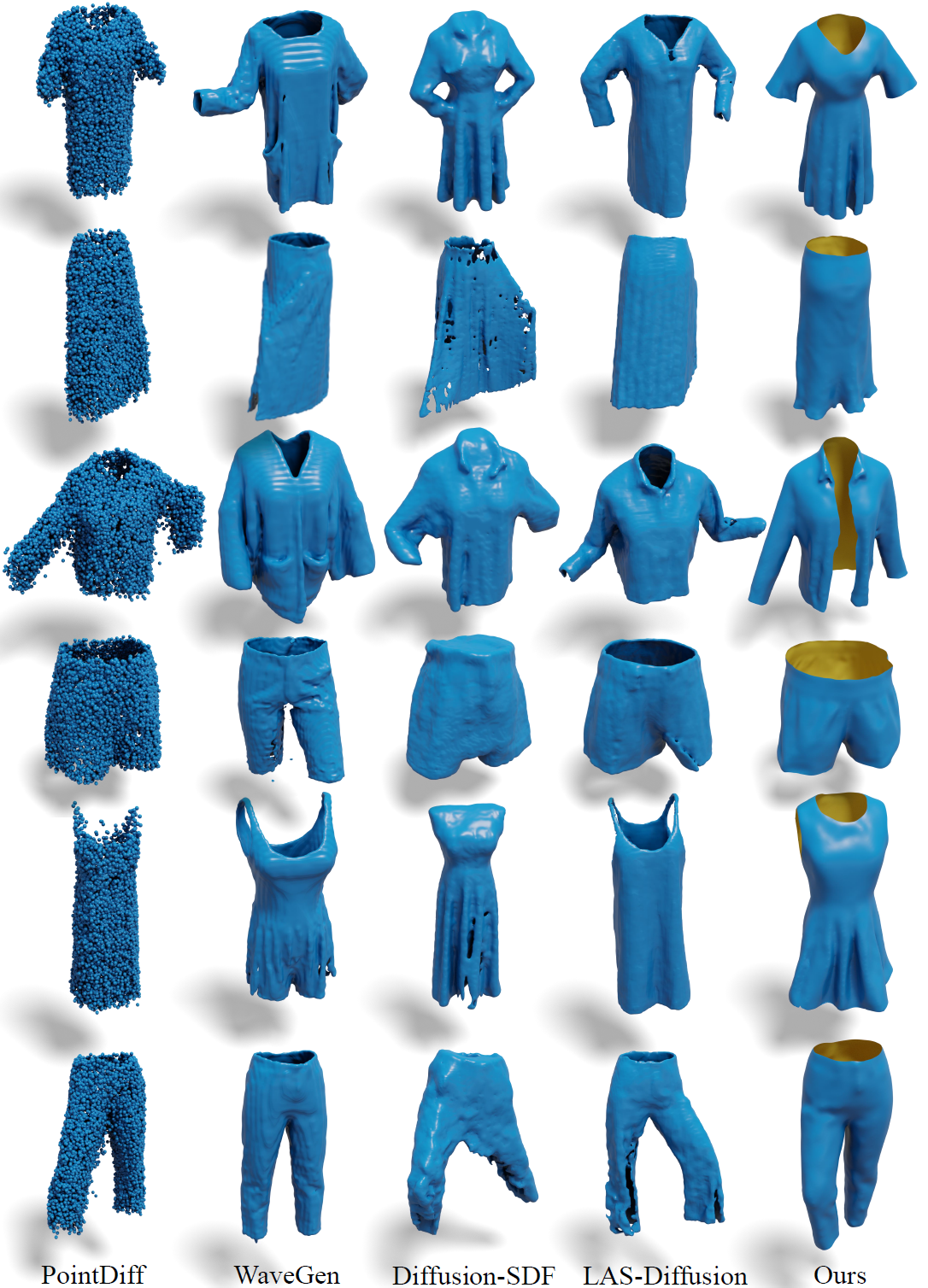} 
  \caption{Visual comparison with state-of-the-arts on the generated shapes under DeepFashion3D dataset. The front and back faces are rendered with different colors  for a clear distinguish on open surfaces. }
  \label{fig:deepfashion}
  \vspace{-0.5cm}
\end{figure}

\begin{figure}[!t]
  \centering
  \includegraphics[width= \linewidth]{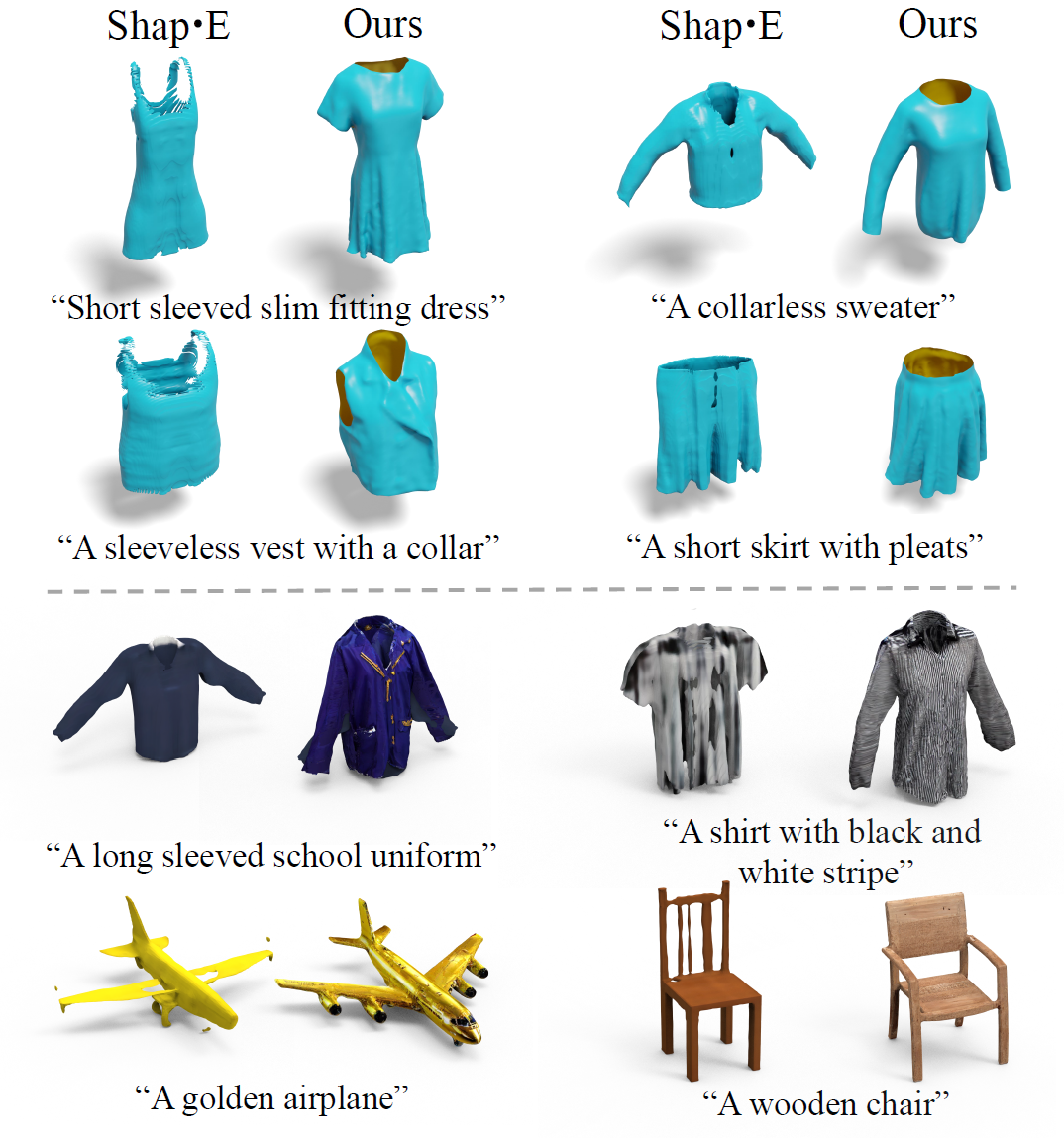} 
  \vspace{-0.7cm}
  \caption{Conditional generations produced by UDiFF and Shap$\cdot$E. The front and back faces are rendered with different colors  for a clear distinguish on open surfaces.  }
  \label{fig:tex_vis}
  \vspace{-0.5cm}
  
\end{figure}

In this section, we evaluate our proposed UDiFF on the task of shape generation. We first demonstrate the performance of UDiFF in generating novel shapes with open surfaces in Sec. \ref{sec.4.1}. Next, we conduct experiments on generating shape with closed surfaces in Sec. \ref{sec.4.2}. The ablation studies are shown in Sec. \ref{sec.4.3}.

\subsection{Open-Surface Shape Generation}
\label{sec.4.1}
\noindent\textbf{Dataset.}
For evaluations in the task of open-surface shape generation, we conduct experiments under DeepFashion3D dataset \cite{zhu2020deep}. The DeepFashin3D dataset is a real-captured 3D dataset of open-surface clothes, containing 1,798 models reconstructed from real garments. It covers 10 categories and 563 garment instances. The dataset is randomly split into training and testing sets by the ratio 80\% and 20 \%. To get the text conditions for training UDiFF, we first render each model from the front facing view to obtain the image representing the model. We then leverage BLIP2 \cite{li2023blip} for captioning the images and keep the caption description of the rendered image for each model as the text condition for the model. We further mix the category description of each model provided in the dataset into the text condition as a supplementary.

\noindent\textbf{Metrics.}
For a fair comparison with various methods, we conduct the quantitative evaluations on the unconditional shape generation. We randomly generate 1,000 shapes with the trained model and uniformly sample 2,048 points on each generated shape. We follow previous works \cite{hui2022neural, luo2021diffusion} to evaluate the generation quality using Minimum matching distance (MMD), Coverage (COV) and 1-NN classifier accuracy (1-NNA). MMD measures the geometry accuracy of the generated shapes. COV indicates the ability of the generated shapes to cover the shapes in the test set. 1-NNA is designed to measure how well a classifier differentiates the generated shapes from the given shapes in the testing set. Lower is better for MMD, higher is better for COV and the closer to 50 \% the better for 1-NNA. 

\noindent\textbf{Baselines.}
We compare UDiFF with the state-of-the-art methods in terms of the shape generation quality. PointDiff \cite{luo2021diffusion}, WaveGen \cite{hui2022neural}, Diffusion-SDF \cite{chou2023diffusion} and LAS-Diffusion \cite{zheng2023locally}. PointDiff uses point cloud data for training, where we sample 2,048 points on each model and leverage the official code for training. All the previous implicit-based shape generation methods represent shapes as SDF or Occ, where the watertight meshes are required to generate the SDF/Occ data for training. Therefore, we leverage the commonly-used manifold method \cite{huang2018robust} for preprocessing the open-surfaces in DeepFashion3D. After that, we follow the official codes of these methods for training unconditional models with the watertight meshes.

\begin{figure}[!t]
  \centering
  \includegraphics[width= \linewidth]{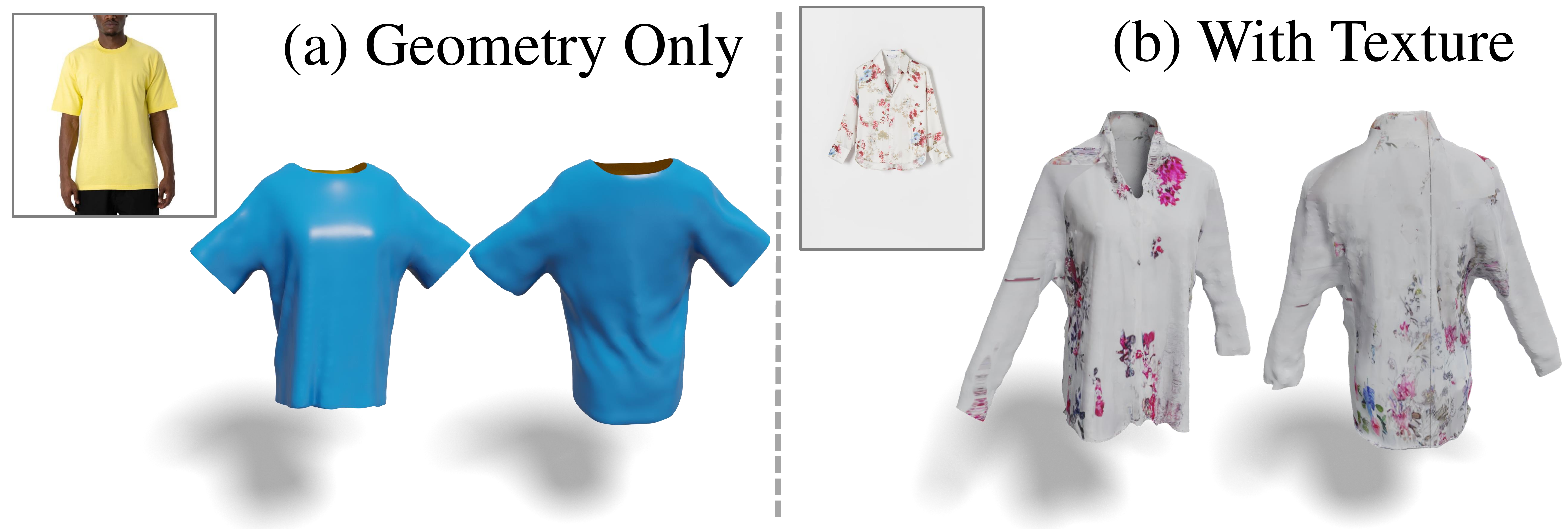} 
  \vspace{-0.6cm}
  
  \caption{Image conditioned generation of UDiFF. (a) Open-surface geometries generated with image guidance. (b) An example of generating textured shapes with image guidance.}
  \label{fig:img_con}
  \vspace{-0.4cm}
\end{figure}

\noindent\textbf{Comparison.}
The quantitative comparison is shown in Tab. \ref{tab.deepfashion}. where UDiFF achieves the best performance compared to  the previous state-of-the-art methods. The main reason is that all the previous implicit-based methods fail to handle the open-surfaces, where the needed manifold preprocessing leads to large bias on the original open-surface shapes. While the proposed UDiFF represents shapes as unsigned distance fields and is able to handle general shapes with or without open surfaces, leading to superior performance compared to other methods. 

The visual comparison is shown in Fig. \ref{fig:deepfashion}, where the proposed UDiFF significantly outperforms the previous works in generating visual-appealing clothes with open surfaces. We render the inside and outside surfaces in different colors for a clear difference on open surfaces.
The PointDiff generates the point cloud to represente shapes, which do not require the manifold preprocess. However, it struggles to produce high-fidelity generations due to the discreteness of points.

\noindent\textbf{Text-conditional Generation.} 
For evaluations in generation with conditions, we further train a conditional model and generate shapes with the guidance from provided text prompts. We visually compare the generations with those produced by Shap$\cdot$E under the same texts as shown in Fig. \ref{fig:tex_vis}. The results demonstrate that UDiFF generates more accurate and high-fidelity predictions from the texts. UDiFF also produces more realistic textures thanks to the powerful Text2Tex \cite{chen2023text2tex}. On the contrary, Shap$\cdot$E struggles to generate correct geometries and textures.

\noindent\textbf{Image-conditional Generation.} 
We further justify that UDiFF can receive diverse signals except texts (e.g. images) for conditional generation. This is achieved by leveraging the pre-aligned text and image representations of the CLIP model, where we adopt the frozen CLIP image encoder to achieve the image embeddings to guide UDiFF generation by cross-attention, without requiring extra training on images. 
We show the image-conditional generations of UDiFF in Fig. \ref{fig:img_con}. The textures on the right of Fig. \ref{fig:img_con} is achieved with Text2Tex \cite{chen2023text2tex} on the text prompt predicted from the image with BLIP2 \cite{li2023blip}, i.e., `A white floral shirt with a long sleeves' .

\begin{table*}[t!]
\centering
\newcolumntype{Y}{>{\centering\arraybackslash}X}
\caption{\textbf{Quantitative comparison of shape generation under ShapeNet dataset.} 
MMD-CD scores and MMD-EMD scores are scaled by $10^3$ and $10^2$, respectively. }
\vspace{-0.2cm}
\footnotesize
{
\setlength{\tabcolsep}{0.2em}
\renewcommand{\arraystretch}{1.0}
\definecolor{LightCyan}{rgb}{0.88,1,1}
\definecolor{Gray}{gray}{0.85}
\begin{tabularx}{\linewidth}{>{\centering}m{3.5cm}| Y Y Y Y Y Y | Y Y Y Y Y Y }
  \toprule
  \multirow{3}{*}{Method} & \multicolumn{6}{c|}{Chair} & \multicolumn{6}{c}{Airplane} \\
   & \multicolumn{2}{c}{COV $\uparrow$} & \multicolumn{2}{c}{MMD $\downarrow$} & \multicolumn{2}{c|}{1-NNA $\downarrow$}  & \multicolumn{2}{c}{COV $\uparrow$} & \multicolumn{2}{c}{MMD $\downarrow$} & \multicolumn{2}{c}{1-NNA $\downarrow$} \\
    &   CD   &   EMD   &   CD   &   EMD   &   CD   &   EMD   &   CD   &   EMD   &   CD   &   EMD   &   CD   &   EMD \\
  \midrule
  IM-GAN  \cite{chen2019learning} & \textbf{56.49}  & 54.50   & 11.79  & 14.52   & 61.98  & 63.45   & 61.55  & 62.79   & 3.320  & 8.371   & 76.21  & 76.08 \\
   Voxel-GAN  \cite{kleineberg2020adversarial}   & 43.95  & 39.45   & 15.18  & 17.32   & 80.27  & 81.16   & 38.44  & 39.18   & 5.937  & 11.69   & 93.14  & 92.77 \\
   PointDiff  \cite{luo2021diffusion} & 51.47  & {55.97}   & 12.79  & 16.12   & 61.76  & 63.72   & 60.19  & 62.30   & 3.543  & 9.519   & 74.60  & 72.31 \\
   SPAGHETTI \cite{hertz2022spaghetti} & 49.48 &	50.22 &	14.7  &	15.85 &	72.34 &	69.46 &	56.86 &	58.83 &	4.260 &	8.930 &	79.36 &	78.86 \\
    SALAD (Global) \cite{koo2023salad}  & 49.71  &	48.75 &	11.71 &	{14.12} &	62.72 &	61.25 & 54.88 &	59.33 &	3.877 &	8.958 &	82.20 &	80.35\\
   SALAD \cite{koo2023salad}  & 56.42	& 55.16	& {11.69}	& 14.29	& \textbf{57.82}	&\textbf{ 58.41}	& {63.16}	& \textbf{65.39}	& 3.636	& 8.238	& \textbf{73.92}	& \textbf{71.08} \\
   WaveGen \cite{hui2022neural}  & 49.63 &	50.15 &	12.12 &	14.25 &	65.04 &	62.87 & 60.94 &	59.09 &	3.528 &	7.964 & 75.77 &	72.93 \\
   \midrule
   Ours & 52.58 & \textbf{55.99} & \textbf{11.67} & \textbf{14.04} & 65.96 & 63.42 & \textbf{64.77} & 63.78 & \textbf{3.151} & \textbf{7.798} & 74.48 & 78.99 \\ 
   
  \bottomrule
\end{tabularx}
}
\vspace{-0.4cm}
\label{tab.shapenet}
\end{table*}

\begin{figure}[!t]
  \centering
  \includegraphics[width= 0.95\linewidth]{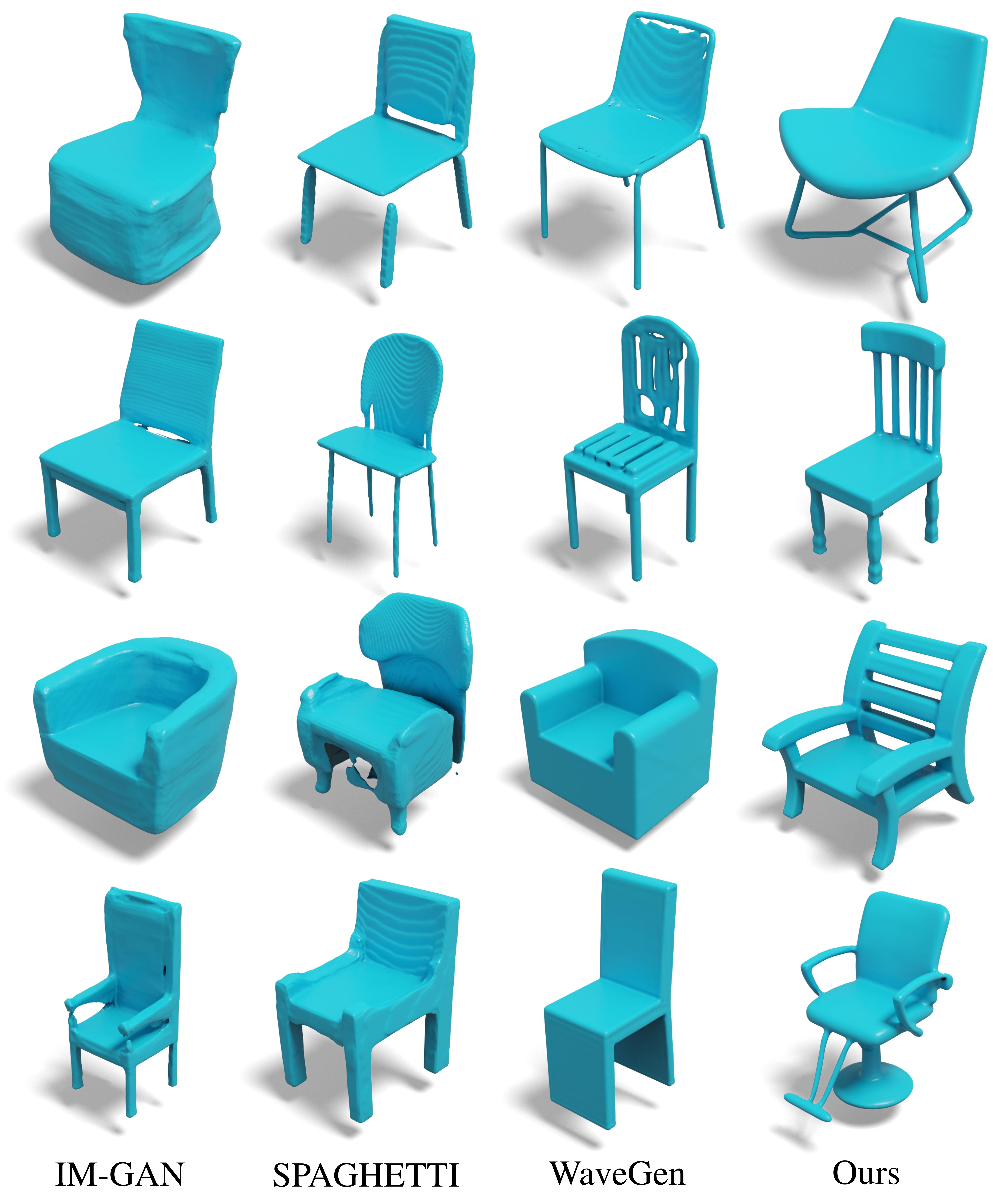} 
  \vspace{-0.3cm}
  \caption{Visual comparison with state-of-the-arts on the generated shapes under ShapeNet dataset.  }
  \label{fig:shapenet}
  \vspace{-0.5cm}
  
\end{figure}

\subsection{Closed Shape Generation}
\label{sec.4.2}

\noindent\textbf{Dataset and metrics.}
For the closed shape generation, we follow the common setting of previous methods \cite{hui2022neural, koo2023salad} to conduct generation experiments under the airplane and chair classes of the ShapeNet \cite{chang2015shapenet} dataset.  We randomly generate 2,000 shapes with the trained model and uniformly sample 2,048 points on each generated shape. We follow previous works \cite{hui2022neural, luo2021diffusion} to evaluate the generation quality using MMD, COV and 1-NNA. We compare our method with all the baselines using their officially provided pretrained models and codes.

\noindent\textbf{Comparison.}
We compare UDiFF with the state-of-the-art methods including IM-GAN \cite{chen2019learning}, Voxel-GAN \cite{kleineberg2020adversarial}, PointDiff \cite{luo2021diffusion}, SPAGHETTI \cite{hertz2022spaghetti}, WaveGen \cite{hui2022neural} and SALAD \cite{koo2023salad}. We show the quantitative comparison in Tab. \ref{tab.shapenet}, where the results are directly borrowed from WaveGen and SALAD for a fair comparison. 

The comparison demonstrates UDiFF also has the capability to generate high-fidelity watertight geometries with only closed surfaces. We justify that UDiFF is a general shape generator to produce general shapes with open surfaces and closed surfaces. We achieve the comparable performance with the state-of-the-art method SALAD, and also significantly outperform the baseline WaveGen which also leverages wavelet transformation as the compact representation. The reason is that our proposed approach for learning optimal wavelet filter largely reduces the information loss during transformation, which leads to more accurate and diverse generations. We further show the visual comparison of some generated shapes of different methods in Fig. \ref{fig:shapenet}. We can see that the shapes generated by our method are more faithful than IM-GAN and SPAGHETTI by producing finer details and cleaner surfaces, and have less bumpy geometries than WaveGen thanks to the optimal wavelet filter to significantly reduce information loss.  

\begin{figure}[!t]
  \centering
  \includegraphics[width= 0.95\linewidth]{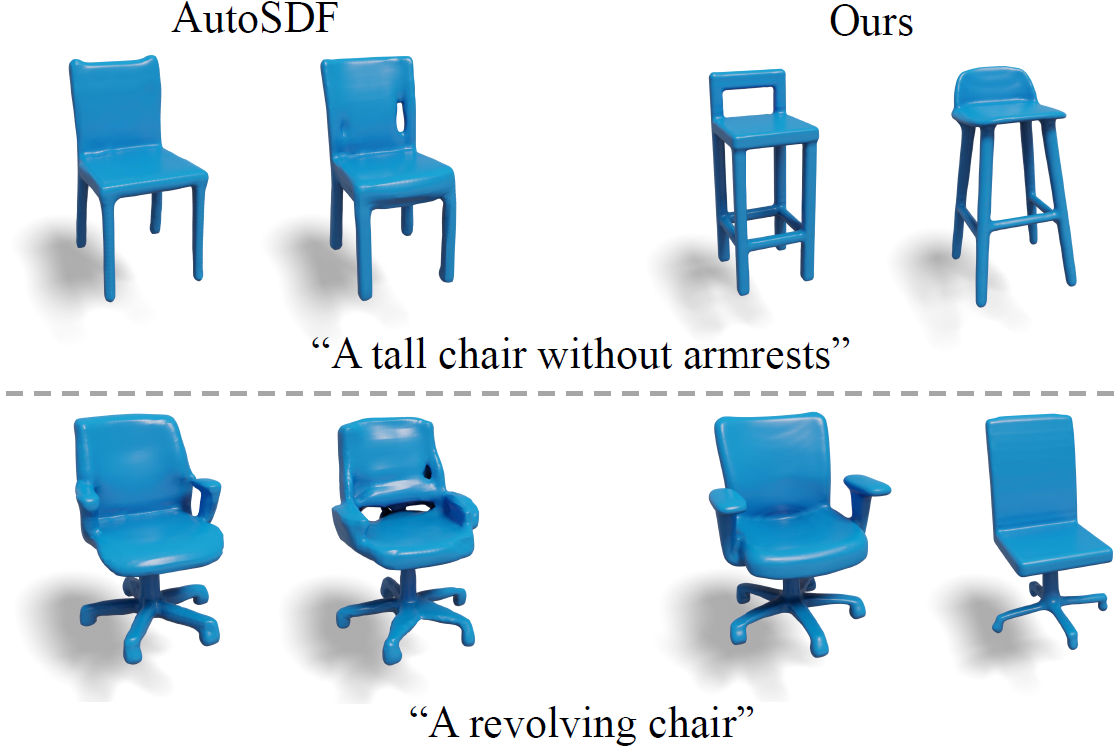} 
  \vspace{-0.3cm}
  \caption{Text-conditioned generation produced by UDiFF and AutoSDF under ShapeNet dataset. }
  \label{fig:con_shapenet}
  \vspace{-0.1cm}
  
\end{figure}

\noindent\textbf{Conditional Generation.} 
We further train a text-conditional model under the `Chair' category of the ShapeNet dataset. We visually compare the generations produced by AutoSDF \cite{mittal2022autosdf} and our propose UDiFF under the same texts as shown in Fig. \ref{fig:con_shapenet}. The results demonstrate that UDiFF generates more accurate and high-fidelity predictions from the texts compared to AutoSDF.

\begin{table}[t!]
\centering
\newcolumntype{Y}{>{\centering\arraybackslash}X}
\caption{\textbf{Ablation studies on the framework design.} MMD-CD scores and MMD-EMD scores are scaled by $10^3$ and $10^2$.}
  \vspace{-0.2cm}

\footnotesize
{
\setlength{\tabcolsep}{0.2em}
\renewcommand{\arraystretch}{1.0}
\definecolor{LightCyan}{rgb}{0.88,1,1}
\definecolor{Gray}{gray}{0.85}
\begin{tabularx}{\linewidth}{>{\centering}m{2.5cm}| Y Y Y Y Y Y }
  \toprule
  \multirow{2}{*}{Method}
   & \multicolumn{2}{c}{COV $\uparrow$} & \multicolumn{2}{c}{MMD $\downarrow$} & \multicolumn{2}{c}{1-NNA $\downarrow$}  \\
    &   CD   &   EMD   &   CD   &   EMD   &   CD   &   EMD   \\
  \midrule
  W/o learned wavelet  & 64.52  & 65.02 & 13.24 & 15.26 & 85.06  & 86.22  \\
  W/o fine predictor  & 66.36 & 65.18 & 12.37  & 14.48 & 83.62 & 84.17  \\
  \midrule
  Full   & \textbf{69.62} & \textbf{67.72} & \textbf{11.60} & \textbf{14.01} & \textbf{81.83} & \textbf{82.14}  \\

  \bottomrule
\end{tabularx}
}
\vspace{-0.3cm}
\label{tab.ablation}
\end{table}

\subsection{Ablation Studies}
\label{sec.4.3}
\noindent\textbf{Framework Design.} To evaluate the major components in our methods, we conduct ablation studies under the DeepFashion3D dataset \cite{zhu2020deep} and report the performance in Tab.~\ref{tab.ablation}. We first justify the effectiveness of the proposed optimal wavelet transformation by replacing our learned wavelet filter with the previous carefully chosen wavelet filter by WaveGen \cite{hui2022neural}, i.e.,  Biorthogonal 6-8. The result is shown as `W/o learned wavelet'. We then remove the fine predictor of UDiFF to recover the 3D shapes with only the generated coarse coefficients as shown in `W/o fine predictor'. The ablation study results demonstrate that effect of designs in UDiFF by significantly improving the generation performance.

\begin{table}[t!]
\centering
\newcolumntype{Y}{>{\centering\arraybackslash}X}
\caption{\textbf{Ablation studies on the effect of wavelet optimization.} We report the L2 Chamfer Distance scaled by $10^5$.}
\vspace{-0.2cm}
\footnotesize
{
\setlength{\tabcolsep}{0.2em}
\renewcommand{\arraystretch}{1.0}
\definecolor{LightCyan}{rgb}{0.88,1,1}
\definecolor{Gray}{gray}{0.85}
\begin{tabularx}{\linewidth}{>{\centering}m{2cm}| Y Y Y  }
  \toprule
  Method & Haar & Biorthogonal3-3 & Biorthogonal6-8 \\
  \midrule
  CD  & 264.8  & 46.04 & 42.92 \\
  \midrule
  \midrule
  Method & Learnable $\phi_{\theta}^D$ & Learnable $\phi_{\delta}^I$ & Both \\
  \midrule
  CD  & 36.12 & 32.15 & \textbf{28.51} \\

  \bottomrule
\end{tabularx}
}
\vspace{-0.45cm}
\label{tab.ablation_wave}
\end{table}

\noindent\textbf{The Effect of Wavelet Optimization.}
We further evaluate the effect of our proposed wavelet optimization to achieve optimal wavelet filter. The result is shown in Tab. \ref{tab.ablation_wave}, where we conduct evaluations under the test set of DeepFashion3D \cite{zhu2020deep} and report the L2 Chamfer Distance between the ground truth meshes and the recovered meshes with wavelet filters Haar, Biorthogonal3-3, Biorthogonal6-8 and ours. We show the performance of only optimizing decomposition filter parameters $\phi_{\theta}^D$ and fix inversion filter parameters $\phi_{\delta}^I$ as `Learnable $\phi_{\theta}^D$', and only optimize $\phi_{\delta}^I$ with fixed $\phi_{\theta}^D$ as `Learnable $\phi_{\delta}^I$'. The best performance is achieved with optimizing both $\phi_{\delta}^I$ and $\phi_{\theta}^D$ as `Both'.

\section{Conclusion}
In this work, we present UDiFF, a 3D diffusion model for conditional or unconditional generating textured 3D shapes with open and closed surfaces. We leverage a diffusion model to learn distributions of UDFs in a spatial-frequency space established through an optimal wavelet transformation for UDFs, which is obtained by data-driven optimizations. The evaluations on widely used benchmarks show our superior performance over the latest methods in generating shapes with either open and closed surfaces.

{
    \small
    \bibliographystyle{ieeenat_fullname}
    \bibliography{main}
}

\newpage

\leftline{\Large{\textbf{Supplementary Material}}}
\renewcommand\thesection{\Alph{section}}
\setcounter{section}{0}

\section{More Visualizations}
In this section, we provide more qualitative illustrations of the generation results produced by UDiFF. 

\begin{figure*}
    \centering
    \includegraphics[width=\linewidth]{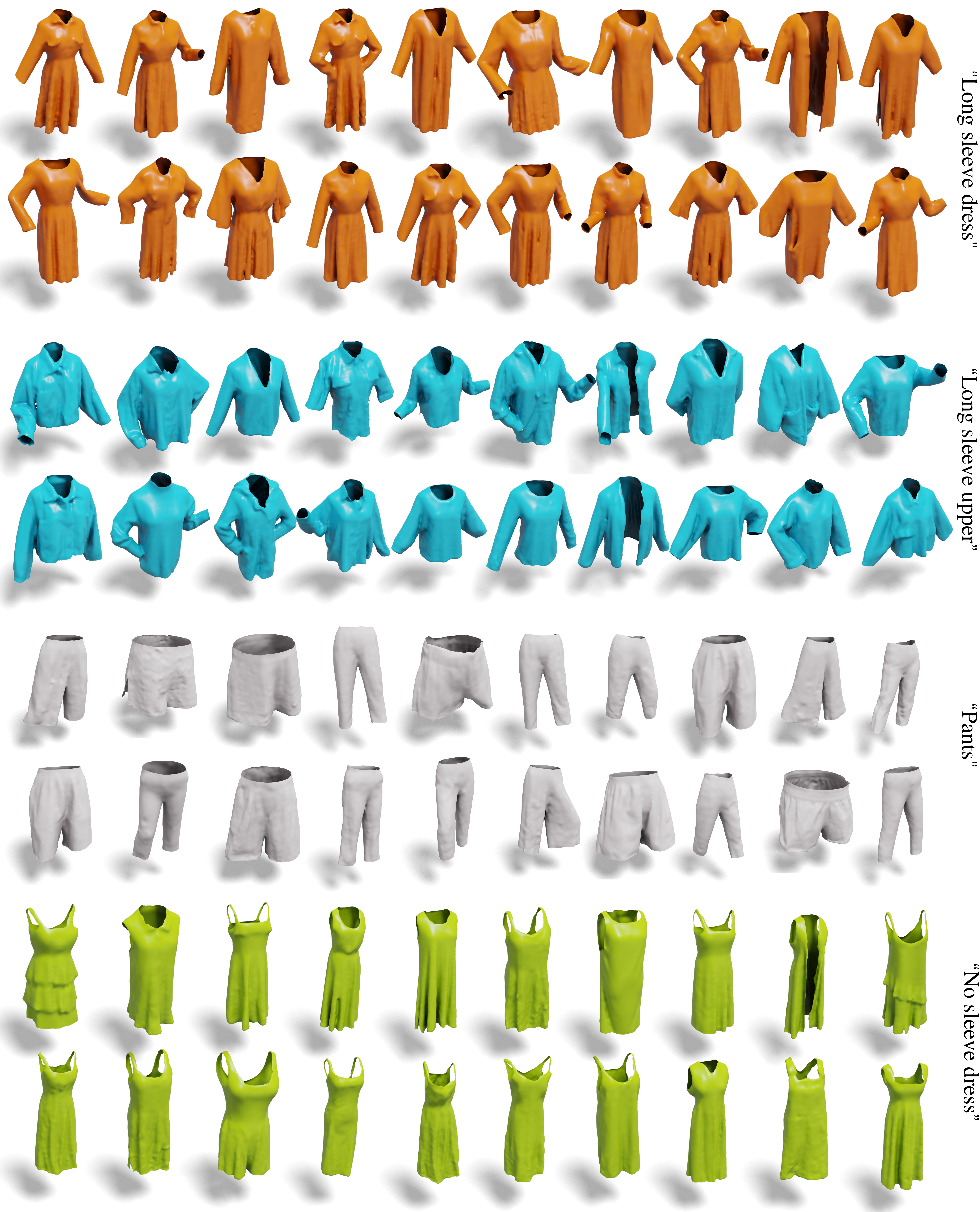}
    \caption{Category conditional generations under DeepFashion3D dataset. Here, we visualize the conditional generations of categories \textit{``long sleeve dress", ``long sleeve upper", ``pants"} and \textit{``no sleeve dress"}}
    \label{fig:deepfashion_supp1}
\end{figure*}

\begin{figure*}
    \centering
    \includegraphics[width=\linewidth]{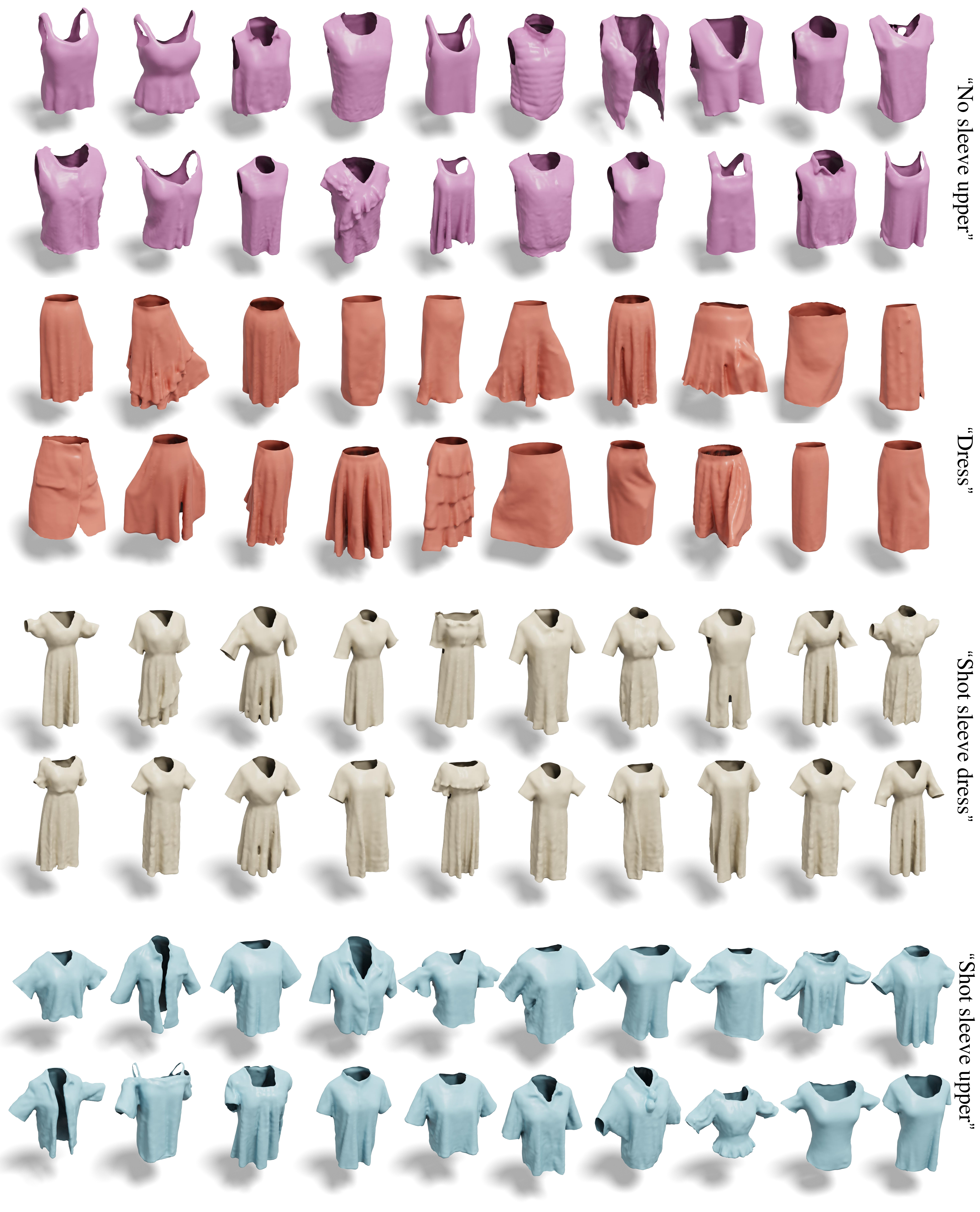}
    \caption{Category conditional generations under DeepFashion3D dataset. Here, we visualize the conditional generations of categories \textit{``no sleeve upper", ``dress", ``shot sleeve dress"} and \textit{``shot sleeve upper"}}
    \label{fig:deepfashion_supp2}
\end{figure*}

\noindent\textbf{Category-conditional generation on DeepFashion3D.} We provide extra open-surface shape generations achieved by the UDiFF model trained under DeepFashion3D \cite{zhu2020deep} dataset with the cloth categories as the conditions. Specifically, we generate 8 categories of cloth shapes, including \textit{``long sleeve dress", ``long sleeve upper", ``pants", ``no sleeve dress", ``no sleeve upper", ``dress", ``shot sleeve dress"} and \textit{``shot sleeve upper"}. The visualizations are shown in Fig. \ref{fig:deepfashion_supp1} and \ref{fig:deepfashion_supp2}, where UDiFF generates diverse and novel shapes correctly corresponds to the text conditions.

\begin{figure*}
    \centering
    \includegraphics[width=\linewidth]{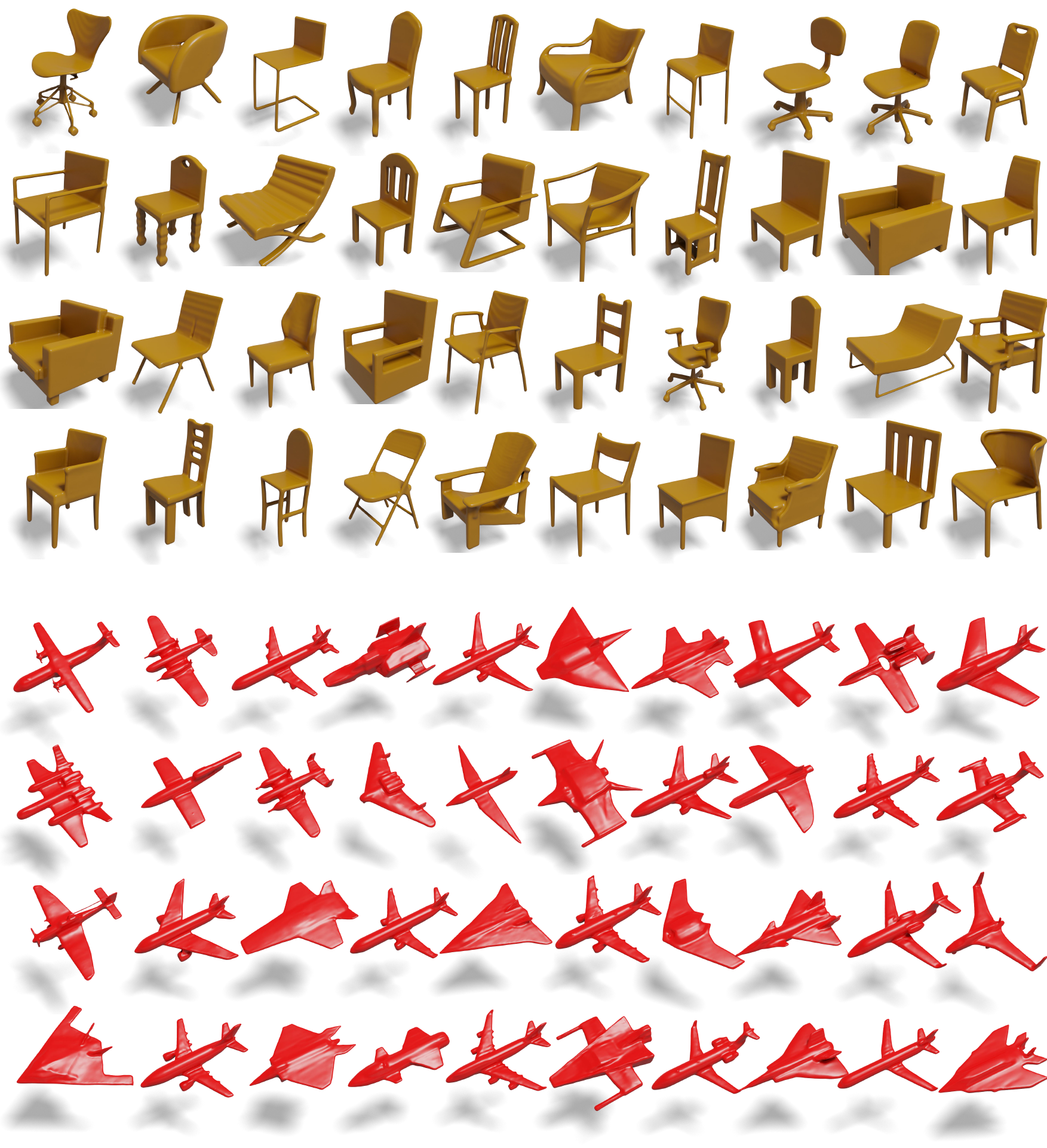}
    \caption{Unconditional generations under the \textit{``chair''} and \textit{``airplane''} categories of the ShapeNet dataset. }
    \label{fig:shapenet_supp}
\end{figure*}

\noindent\textbf{Unconditional generation on ShapeNet.} 
We further provide more unconditional shape generation results achieved by the UDiFF model trained under single categories of ShapeNet \cite{chang2015shapenet} dataset. We generate shapes of the \textit{``chair''} and \textit{``airpane''} categories. The visualizations are shown in Fig. \ref{fig:shapenet_supp}, where UDiFF generates visual-appealing shapes.

\section{Analysis on Meshing and Texturing}
\begin{figure}[h]
    \centering
    \includegraphics[width=0.9\linewidth]{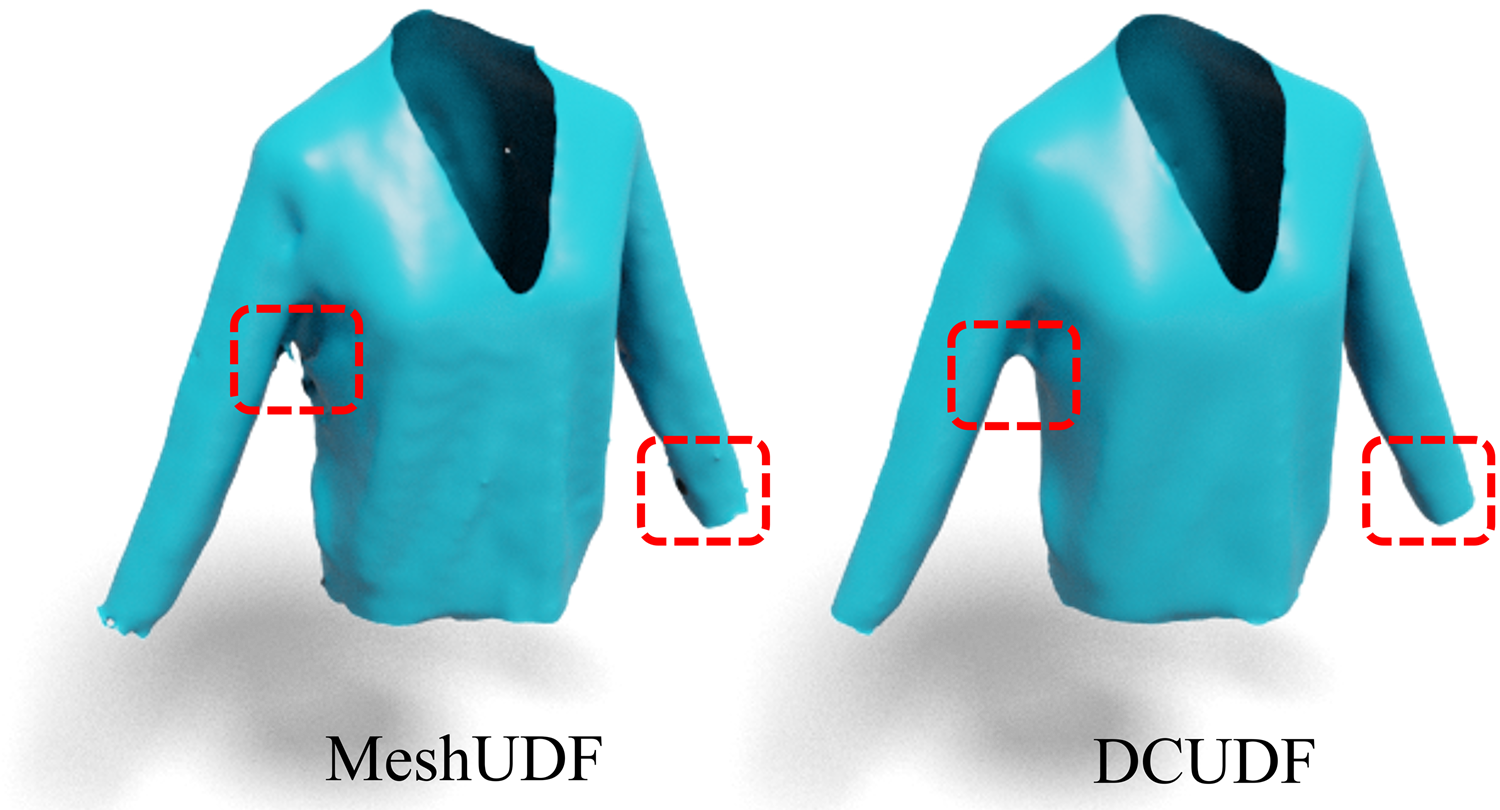}
    
    \caption{Mesh extraction comparisons between MeshUDF and DCUDF.}
    \label{fig:meshing}
\end{figure}

\noindent\textbf{Meshing.} Different from SDFs, UDFs fail to extract surfaces by the marching cubes \cite{lorensen1987marching} since UDFs cannot perform inside/outside tests on 3D grids. Recent works \cite{guillard2021meshudf, Zhou2022CAP-UDF} leverage the gradients at UDF grids as the signals to mesh UDFs. However, for the generated UDFs, the approximated gradients may not be stable enough at the zero-level set, which leads to errors and holes. The approximated gradient at a grid point $q$ is defined as the direction from $q$ to the neighbour grid $q_n$ where the UDF from $q$ to $q_n$ increases rapidly the most. We adopt DCUDF \cite{hou2023robust} with double covering to mesh the generated UDF of UDiFF, which results in more continuous surfaces. We make an adaption to DCUDF on the double covering operation to replace the time-consuming optimizations with an explicit vertices refinement strategy. We move each vertices against the surface normals with a stride of unsigned distances to reach the zero-level sets, and then leverage the min-cut algorithm to achieve the final model.  
We show the comparison of meshing the generated UDF with MeshUDF \cite{guillard2021meshudf} and DCUDF \cite{hou2023robust} in Fig.~\ref{fig:meshing}.

\noindent\textbf{Texturing.} We leverage Text2Tex \cite{chen2023text2tex} to generate textures for the extracted meshes. This is achieved with a progressive texture generation process and a texture refinement process. Specifically, we first render the texture-less initial mesh from the preset viewpoint and generate the appearance according to the text prompt with the depth-guided stable-diffusion \cite{rombach2022high}. We then adjust to the next preset viewpoint and repeat the appearance generation process until the last preset viewpoint where the whole mesh is textured. Finally, we optimize the textures with automatically selected viewpoints for refinement.

\end{document}